\documentclass[runningheads]{llncs}

\usepackage[T1]{fontenc}
\usepackage{devanagari} 
\usepackage{graphicx}
\usepackage{amsmath}
\usepackage{amssymb}
\usepackage{pifont}
\usepackage{graphicx}
\usepackage{tabularx}
\usepackage{booktabs}
\usepackage{enumerate}
\usepackage{mathtools}
\usepackage{color}
\usepackage{url}
\usepackage{amssymb}
\usepackage{multirow}
\usepackage{cite}
\usepackage[table]{xcolor}

\usepackage{color}

\makeatletter
\newcommand{\printfnsymbol}[1]{%
  \textsuperscript{\@fnsymbol{#1}}%
}
\makeatother

\newcommand{\data}{\textsc{VT-Syn}}
\newcommand{\realdata}{\textsc{VT-Real}}

\newcommand{\xmark}{\ding{55}}%

\def\spacing{3pt}
\renewcommand\textfloatsep{\spacing}

\renewcommand\intextsep{\spacing}

\begin{document}
\title{Show Me the World in My Language: Establishing the First Baseline for \\Scene-Text to Scene-Text Translation}
\titlerunning{Show Me the World in My Language}

\author{Shreyas Vaidya\inst{1}\thanks{Equal Contribution.} \and
Arvind Kumar Sharma\inst{1}\printfnsymbol{1}  \and\\
Prajwal Gatti\inst{2}\thanks{This work was done while Prajwal Gatti was affiliated with IIT Jodhpur.}\and
Anand Mishra\inst{1}}

\institute{Indian Institute of Technology Jodhpur \and
University of Bristol\\
\email{\{vaidya.2, sharma.126, mishra\}@iitj.ac.in }\\
\email{prajwal.gatti@bristol.ac.uk}}

\authorrunning{S. Vaidya et al.}

\maketitle
\begin{abstract}
In this work, we study the task of ``visually'' translating scene text from a source language (e.g., Hindi) to a target language (e.g., English). Visual translation involves not just the recognition and translation of scene text but also the generation of the translated image that preserves visual features of the source scene text, such as font, size, and background. There are several challenges associated with this task, such as translation with limited context, deciding between translation and transliteration, accommodating varying text lengths within fixed spatial boundaries, and preserving the font and background styles of the source scene text in the target language. To address this problem, we make the following contributions: (i) We study visual translation as a standalone problem for the first time in the literature. (ii) We present a cascaded framework for visual translation that combines state-of-the-art modules for scene text recognition, machine translation, and scene text synthesis as a baseline for the task. (iii)  We propose a set of task-specific design enhancements to design a variant of the baseline to obtain performance improvements. (iv) Currently, the existing related literature lacks any comprehensive performance evaluation for this novel task. To fill this gap, we introduce several automatic and user-assisted evaluation metrics designed explicitly for evaluating visual translation. Further, we evaluate presented baselines for translating scene text between Hindi and English. Our experiments demonstrate that although we can effectively perform visual translation over a large collection of scene text images, the presented baseline only partially addresses challenges posed by visual translation tasks. We firmly believe that this new task and the limitations of existing models, as reported in this paper, should encourage further research in visual translation. We have publicly released the code and dataset on our project website: \url{https://vl2g.github.io/projects/visTrans/}.
\end{abstract}

\begin{figure*}[!t]
\centering
 \includegraphics[width=\columnwidth]{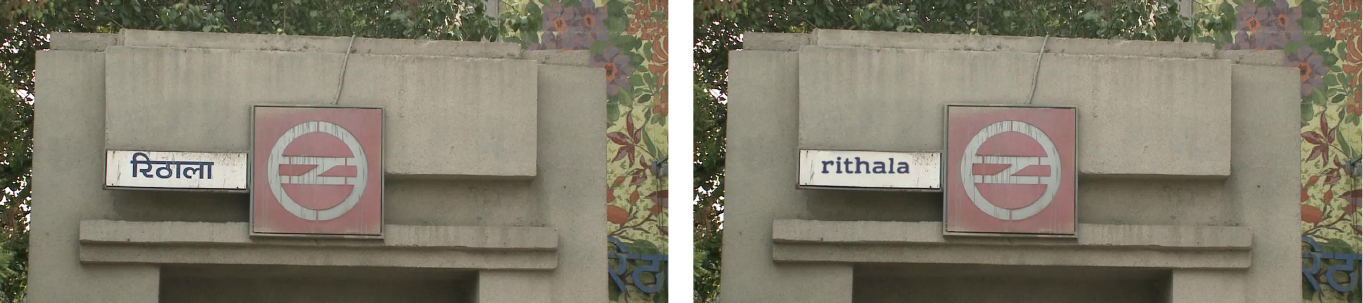}
\caption{Imagine visiting Delhi, India, and arriving at the Rithala (Hindi: {\dn ErWAlA}) metro station. If you are not familiar with Hindi, the signboard on the left might be incomprehensible. The result of our proposed baseline solution, shown on the right, seamlessly transliterates the station name {\dn ErWAlA} to English. In our work, we aim to visually translate (or transliterate, when necessary, as in this case) text from the source language to the target language while preserving the visual attributes of the source scene text. Specifically, we focus on visual translation between Hindi and English in this work.}
\label{fig: goal}
\end{figure*}

\section{Introduction}
\label{sec:intro}
Machine Translation has shown remarkable growth in the last few years, partly attributed to the adoption of neural models~\cite{BahdanauCB14,vaswani2017attention,edunov2018understanding,zhao2019muse,fan2021beyond}. In parallel, substantial advancements have also been made in speech-to-speech translation~\cite{nakamura2006atr,jia2019direct,lee2022direct,kano2021transformer} where the goal is to develop systems that are capable of accurately interpreting spoken language in one dialect and seamlessly translating it into another while preserving the voice of the original speaker, thus enabling effective cross-lingual communication in real-time. Drawing inspiration from these research directions, we present an analogous problem in the scene text domain, namely ``scene-text to scene-text translation'' or, in short, ``visual translation''. The visual translation task aims
to translate text present in images from a source language to the target language while preserving the visual characteristics of the text and background as illustrated in Fig.~\ref{fig: goal}. Visual Translation has extensive applications, e.g., transforming the travel experience by allowing tourists to instantly understand sign boards in foreign languages and enabling seamless interaction with the visual world without language barriers. 

By drawing parallels with the speech-to-speech translation approaches, which comprise three components: automatic speech recognition (ASR), text-to-text machine translation (MT), and text-to-speech (TTS) synthesis, we propose a visual translation baseline that integrates scene-text recognition (STR), text-to-text machine translation (MT), and scene-text synthesis (STS). This cascaded system offers practical advantages over an end-to-end approach, as fully supervised end-to-end training necessitates a substantial collection of source and target scene text pairs, which can be challenging to obtain compared to parallel text pairs for MT or image-text pairs for STR. As STR and MT models are extensively explored in the literature, and several off-the-shelf methods are available, we prioritize enhancing the performance of the STS model. To this end, we extend a popular SRNet architecture~\cite{wu2019editing} by decoupling background and foreground generation. For background generation, we employ a diffusion-based model using ControlNet~\cite{controlNet} to generate a text-erased image from an input containing scene texts. We further modify SRNet so that it only focuses on foreground generation on a plain background. Once foreground and background are independently generated, we blend them into the scene image. To improve the quality of visual translation further, we propose a set of design enhancements such as using regular expressions to filter special strings, grouping words and translating them together, and a planning strategy to blend the translated text in the scene image appropriately.

We extensively evaluate the proposed baselines for Hindi-to-English and English-to-Hindi visual translation using our new automatic and user evaluation metrics. While the baselines show promising results, the problem remains far from solved and requires further research.

We make the following contributions: (i) We study the under-explored task of visual translation that aims to translate text in images to a target language while preserving its font, style, position, and background. To the best of our knowledge, the comprehensive study of this problem has largely been unexplored in the existing literature. (ii) We introduce a generic cascaded approach for visual translation, and we design a set of baselines using state-of-the-art approaches for scene text recognition, machine translation, and scene text synthesis and their task-specific design enhancements. (iii) Training a visual translation model with real-world images is challenging due to the lack of large-scale paired scene text images in different languages. Therefore, we use synthetic images for training. We present a method to generate paired images with words sharing the same visual properties, creating \data{}, a synthetic dataset of 600K paired visually diverse English-Hindi scene-text images. To evaluate performance on real images, we provide extensive annotations of translated text from three users. These benchmark datasets will support future research in visual translation. (iv) Due to the lack of principled evaluation metrics for visual translation tasks in the literature, we propose a set of automatic and user evaluation metrics. We believe these metrics will help track the progress of visual translation tasks effectively.
\section{Related Work}
\textbf{Machine Translation:} It is a well-studied area~\cite{desai2014domain,hurskainen2017rule,pirinen2019apertium,gehring2017convolutional,vaswani2017attention,edunov2018understanding} that aims to convert a text from its source language to a target language. Current state-of-the-art models for machine translation are deep-learning based~\cite{gehring2017convolutional,vaswani2017attention,edunov2018understanding}. In the speech domain, Speech-to-Speech Translation (S2ST) aims to translate speech from one language to another while preserving the speaker's voice and accent~\cite{jia2019direct,lee2022direct,kano2021transformer}. Inspired by these works, we focus on text translation in the visual modality, which brings newer research challenges, such as preserving font properties and integrity of the image background, which need to be addressed to produce visually appealing translations.
\newline
\newline
\noindent\textbf{Translation of Text in Images:} Recent years have seen growing interest in translating text within images, both in research and commercial domains. Current works primarily focus on recognition and translation methods for scene text~\cite{ma2022improving,lan2023exploring}, prioritizing accurate translation without addressing visually consistent text generation. A popular commercial product -- Google lens~\footnote{\url{https://lens.google/#translate}} also falls into this category. These approaches often resort to simply overlaying translated text on source images. While some studies explore end-to-end methods for text translation in images that generate text directly in pixel space~\cite{mansimov2020towards,lan2024translatotron}, they typically deal with limited visual diversity in document-style images with plain backgrounds and fixed fonts – without tackling the complexities of scene text that we aim to address. The closest solution to our problem is Google Translate for images\footnote{\url{https://translate.google.com/?op=images}}, a commercial product for visual translation of diverse scene text. However, its underlying technology remains proprietary and closed-source. We emphasize the need for the research community to study this problem openly, establish proper open-source solutions, create a benchmark, and define evaluation criteria – goals we pursue in this paper. Moreover, we observe that Google Translate still lacks translation quality and often fails to produce visually consistent results for complex cases, underlining the potential for better approaches. In Section~\ref{sec:comm}, we provide a qualitative comparison between our work and Google Translate.
\label{sec:method}
\begin{figure}[!t]
    \centering
    \includegraphics[width=\columnwidth]{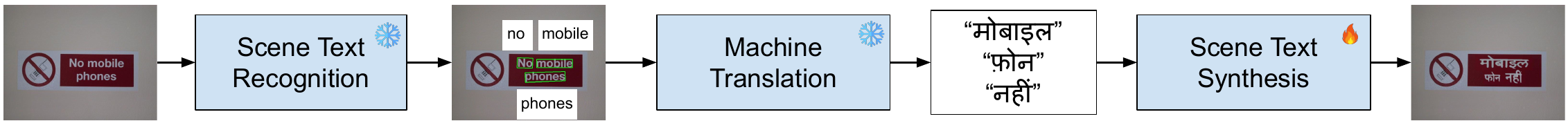}
    \caption{\textbf{Outline of proposed cascaded baseline for Visual Translation.} We use state-of-the-art approaches for scene text recognition, machine translation, and scene text synthesis to design variants of our baseline. Moreover, we further investigate the scene-text synthesis and propose an extension to existing SRNet architecture.}
    \label{fig:outline}
\end{figure}
\newline
\newline
\noindent\textbf{Editing Text in Images:}
The problem of editing text in images has witnessed significant research interest in recent years~\cite{wu2019editing,yang2020swaptext,roy2020stefann,lee2021rewritenet,krishnan2023textstylebrush,qu2023exploring}. This task aims to modify scene text to target content while retaining visual properties of the original text. SRNet~\cite{wu2019editing} is one such method that learns an end-to-end GAN-based style-retention network. SwapText~\cite{yang2020swaptext} improved upon the SRNet architecture by modeling the geometrical transformation of the text using spatial points. More recently, TextStyleBrush~\cite{krishnan2023textstylebrush}, RewriteNet~\cite{lee2021rewritenet}, and MOSTEL~\cite{qu2023exploring} introduce a self-supervised training approach on real-world images for this task. Further, TextStyleBrush is evaluated on handwritten images as well. Authors in~\cite{roy2020stefann} proposed a character-wise text editor model for this task. However, their approach assumes source and target text instances are of the same length, which is not always true, especially in the translation task. A more recent approach, MOSTEL~\cite{qu2023exploring}, also introduces stroke-level modifications to produce more photo-realistic generations. Despite these advances, these methods only address the cross-lingual editing problem, which is just a component of the visual translation process and is insufficient on its own for achieving visual translation. Our work aims to address the task of visual translation and its complexities more comprehensively.
\section{Proposed Visual Translation Baseline}

\begin{figure*}[!t]
\centering
 \includegraphics[width=0.9\textwidth]{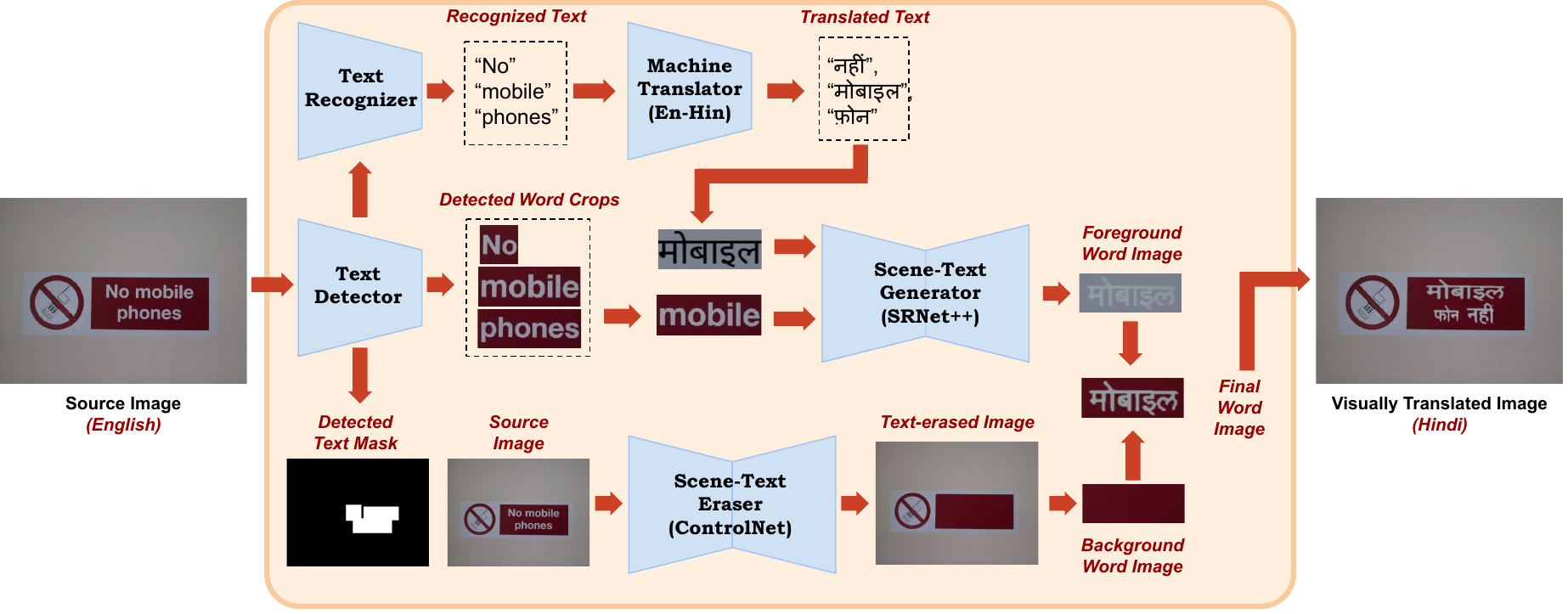}
\caption{\textbf{Our proposed baseline} extends the SRNet scene text synthesis approach by decoupling background and foreground generation. More details provided in Section~\ref{sec:method}.}
\label{fig: method}
\end{figure*}

The task of Visual Translation can be reduced to a sequence of sub-tasks: locating and reading text in scene images, translating the text into the target language, and generating the final image containing the translated scene text. Motivated by this observation, we propose a cascaded approach to visual translation by combining models for (i) scene text recognition, (ii) machine translation, (iii) scene-text synthesis, and (iv) seamlessly blending the generated scene text into the image. These sub-tasks are well-explored independently in computer vision literature; thus, we benefit from the availability of trained models. Further, such an approach can perform generation at the word or phrase level, which can help preserve the consistency of non-text regions in the image. The outline of our cascaded baseline is illustrated in Fig.~\ref{fig:outline} and a detailed illustration is provided in Fig.~\ref{fig: method}. We describe each module in detail in the following sections.

Training and evaluation of a visual translation baseline require real-world images in the form of $(I, I')$ where $I$ and $I'$ are visually identical images containing corresponding scene text in two different languages with matching font and style. However, such instances are not easily available in the real world. We mitigate this data scarcity challenge by directly \textit{synthesizing} the desired data: we generate paired scene-text images that are (i) identical in the image background and (ii) matching in font and style. A few examples are shown in Fig.~\ref{fig:dataset-examples}. 

In generating synthetic samples, we use a large corpus of words in both languages, as well as a diverse collection of fonts. To simulate real-world scene text, we also render the images on natural backgrounds, as well as vary the orientation, positioning, and size of the scene text in images. A more detailed procedure for generating the synthetic data is provided in Section~\ref{sec: synthetic-data}.

\noindent\textbf{(i) Locating and Recognizing Text in Images.}
The first step in our proposed baseline is locating scene text in images, followed by recognizing the detected text, which are both well-explored problems in computer vision literature. Given the source image, we use a scene-text detector to detect all occurrences of text in the image by predicting a bounding box around them. Next, we use a text recognition model that predicts the text content from the crops of words obtained from the previous step. In this work, we use DBNet~\cite{liao2020real} for text detection and ParSeq~\cite{bautista2022scene} for the text recognition step pretrained on English and Hindi language data, respectively.

\noindent\textbf{(ii) Machine Translation of Text.} After obtaining the recognized text in the source language $L$, we map each instance to the desired target language $L'$ using an off-the-shelf neural machine translation method. We test our model with two state-of-the-art neural machine translation methods, namely IndicTrans2~\cite{gala2023indictrans2} and M2M100~\cite{fan2021beyond}. IndicTrans2 is trained on a large collection of Indic languages (including Hindi), whereas M2M100 is a more general translation method trained on a diverse collection of languages with support for Indic languages as well.

\noindent\textbf{(iii) Scene-Text Synthesis.}
Our pipeline has thus far obtained source word bounding boxes, recognized text, and translated text. The final step is to generate the target word image containing the translated text while maintaining stylistic consistency with the source text.

Popularly, scene-text synthesis methods aim to generate this image using a single end-to-end trained model~\cite{wu2019editing,qu2023exploring,krishnan2023textstylebrush,lee2021rewritenet}. However, in our empirical results, we consistently observed two frequent limitations with such methods: (i) incomplete erasure of source text in the generated image and (ii) unintended alterations to the background (non-text) regions. These errors often resulted in a "patchy" effect, among other unnatural artifacts, when the generated images were integrated into the full scene.

To address these issues, we propose a decoupled approach to scene-text synthesis, consisting of three independent steps: background generation and foreground generation, followed by a composition step. We collectively term this approach SRNet++ -- an enhancement to the original SRNet architecture~\cite{wu2019editing}. The proposed SRNet++ works as follows: (a) \textbf{Background Generation}: In this step, we employ a diffusion-based model using the ControlNet architecture to generate a text-erased image from an input containing scene texts using a publicly available implementation~\cite{DiffSceneTextEraser}. The model is conditioned on a binary text-masked image. It takes the full-sized source image and a mask image indicating detected text regions as input, producing a full-size image with text regions erased. These text-erased regions are then cropped to obtain clean background images.
(b) \textbf{Foreground Generation}: In this, we modify the SRNet~\cite{wu2019editing} architecture to generate only the foreground text information on a plain background. The model takes a source word crop and target text (rendered as black text on a gray background) as input, generating colored foreground scene text on a gray as output. During training, the model is optimized to generate both the target text and a skeletal image of the text. This model is trained from scratch using synthetic data. 
(c) \textbf{Composition Step}: This step combines the generated background and foreground images for each word. We apply Otsu's method~\cite{otsu1975threshold} to the foreground image to obtain a thresholded binary mask, which is used to extract the foreground text region. The extracted text is then composited onto the background image. This approach results in a clear, smooth image that maintains visual consistency with the source, avoiding the jagged-edge artifacts that can occur with simple overlay methods. The background generation model utilizes pre-trained weights, while the foreground generation model is trained from scratch on our synthetic dataset. In Section~\ref{sec:experiments}, we compare the decoupled approach of our proposed SRNet++ with direct scene-text generation methods of MOSTEL~\cite{qu2023exploring} and SRNet~\cite{wu2019editing}.

Once we obtain all the target word images through this process, we compose them onto the full-sized input image at their respective positions. This final composition step yields the complete visually translated image.

\subsection{Design Enhancements}
\label{subsec:design}
To enhance the design, a series of refined and newly introduced steps have been implemented. The process begins with the detection and recognition of text, after which numbers, websites, and email addresses are filtered out using regular expressions. Note that these elements do not need to be translated. Words are then grouped into paragraphs and lines based on the geometry and coordinates of the bounding box in conjunction with a heuristic function. These paragraphs are translated and segmented into lines, ensuring alignment with the proportion of lines present in each original paragraph. Through cubic spline interpolation, new coordinates for each word within a line are determined, which are then linked back to the original crops of the words. Depending on the new width of the translated words, adjustments are made to the crops—either cutting or replicating them—to maintain the original style of the text. 

The process is finalized by accurately positioning the new words on the image using the developed method. Although heuristically designed, this step shows a significant boost in translation quality, as shown in the experiments. 
\subsection{Baseline Variants}
\label{subsec:baselines}
We present several baseline variants for visual translation, each incorporating different combinations of techniques for scene text detection, recognition, machine translation, and image synthesis. These variants are designed to evaluate the impact of each individual component and improvements in the pipeline. \textbf{B-1:} Utilizes ground truth scene text detection and recognition, pre-trained M2M100~\cite{fan2021beyond} for machine translation, and SRNet~\cite{wu2019editing} for scene text synthesis. \textbf{B-2:} Identical to B-1, but uses MOSTEL~\cite{qu2023exploring} instead of SRNet for scene text synthesis. \textbf{B-3:} Modifies B-1 by employing SRNet++ (our proposed enhancement of SRNet) for scene text synthesis. \textbf{B-4:} Modifies B-3 by replacing oracle bounding boxes with state-of-the-art DBNet~\cite{liao2020real} for detection and ParSeq~\cite{bautista2022scene} for recognition. \textbf{B-5:} Modifies B-3 by substituting M2M100 with IndicTrans2~\cite{gala2023indictrans2}, a state-of-the-art translation module for Indic languages. \textbf{B-6:} Identical to B-5 but uses DBNet and ParSeq instead of using Oracle bounding boxes. \textbf{B-7:} Addresses the limitations of word-level translation by incorporating the design enhancements proposed in Section~\ref{subsec:design}. This variant is built upon the best-performing baseline from B-1 to B-6. B-7, in particular, represents a significant departure from the word-by-word translation approach, accounting for language-specific word ordering and context. 

\begin{figure}[!t]
\centering
\includegraphics{Figures/Fig4_dataset.pdf}
\caption{\data{} dataset examples, which contains paired Eng $\rightarrow$ Hin and Hin $\rightarrow$ Eng images with diverse fonts, text colors, sizes, orientations, and background images of natural scenes, textures, and plain colors.}
\label{fig:dataset-examples}
\end{figure}

\section{Dataset}
\label{sec:data}
The problem of visual translation has not been comprehensively studied in the literature. Therefore, no benchmark dataset currently exists for its comprehensive investigation. To fill this gap, we present the following datasets: 
\subsection{\data{}: Synthetic Training Data}
\label{sec: synthetic-data}
For training the scene text synthesis components of our pipeline, we need paired images of text in different
languages with identical visual properties (style, font, orientation, and size.). It is extremely difficult to get visually identical scene images with text in different languages in the real world, and it is even more difficult to generate accurate skeleton images required for training SRNet, MOSTEL, as well as our proposed SRNet++ method. Thus, we rely on generating highly diverse synthetic images. We introduce \textbf{\data{}}, a synthetically generated corpus of 600K visually diverse paired bilingual word images in pairs of English-Hindi as well as Hindi-English.

We utilized an Indic-language scene-text image generator~\cite{mathew2017benchmarking} and modified it to generate samples of scene-text in paired languages with controllable parameters for font, style, color, and spatial transformations to ensure visual diversity. Each sample contains a source image, a target word image, a background image, a foreground image, and a target image. We also generate source word images, source and target masks, and skeleton images based on the requirements of various scene-text synthesis architectures. We collect 291 publicly available fonts that support both Roman and Devanagari scripts and use a vocabulary of 3K commonly used words in both languages. 

A few samples from \data{} are shown in Fig.~\ref{fig:dataset-examples}. Note that the paired image words do not have to be translations as the STS module has to particularly learn to render the target word using the same style as the source image.

\subsection{\realdata: Real Test Dataset}
For the purpose of evaluation, we propose \realdata, which contains images from 
ICDAR 2013~\cite{Karatzas2013ICDAR2R} 
and Bharat Scene Text Dataset~\cite{BSTD} to evaluate English-to-Hindi and Hindi-to-English translations, respectively. We filter images of moderate complexity\footnote{As a first work on Hindi-to-English and English-to-Hindi translation, we have opted not to include highly complex curved and occluded text.} from these two sources. In all, our dataset contains 269 images and 1021 words. These images were given to three human annotators to translate the text from Hindi to English and vice versa. A few example translation annotations of this dataset are shown in Fig.~\ref{fig:translations}. Even though the above-mentioned datasets have no ground truths for Visual Translation (i.e., scene text in the target language), they are still useful for automatic evaluation proposed in the next section. 

\begin{figure}[!t]
\centering
\includegraphics[width=\columnwidth]{Figures/Fig5_vtreal_translations.pdf}
\caption{A few examples from \emph{VT-Real} dataset, showing image and Eng-Hin and Hin-Eng ground truth translations, manually annotated by three independent annotators (referred to as users here).}
\label{fig:translations}
\end{figure}

\section{Performance Metrics}
\label{sec:pm}
Evaluating visual translation methods is complex, even more so than evaluating machine translation. While the evaluation of machine translation has been a longstanding research area in the NLP community, recent research has saturated the use of metrics such as BLEU, METOER, and ROUGE; visual translation poses several additional challenges. Unlike machine translation, which is typically evaluated for a sentence or paragraph of text, in visual translation, one has to evaluate the correctness of translation for a single word or a small set of words or phrases. Further, It requires not only assessing the linguistic accuracy of the translation but also ensuring the preservation of background and font properties.

In this work, we propose automatic and user evaluation as follows:
\subsection{Automatic Evaluation}
\label{sec:autoEval}
We proposed the following three automatic evaluation metrics:

\noindent\textbf{(i) Translation Quality (TQ)}: To measure translation quality, we first detect and recognize scene text in the target language. We then group them and send them to an off-the-self machine translator, i.e., IndicTrans2. We then evaluate BLEU-1 and BLEU-2 to measure translation quality using reference translation annotations for each image and report mean scores for all images. It should be noted that BLEU-2 is not computed for those images where there is only one word in the target translation. Note that all BLEU scores are computed along with smoothing techniques as suggested in literature \cite{chen2014systematic}

\noindent\textbf{(ii) Perception Quality (PQ)}: Visually translating also requires the model to generate perceptually high-quality images without any patches or artifacts. To evaluate perception quality, we propose to use CONTRastive Image QUality Evaluator, {\sc contrique}~\cite{Contrique}, a recent approach for image quality assessment without any reference. 

\noindent\textbf{(iii) $\mathbf{VT\text{-}score}$}: For high-quality visual translation, it is important to have a high BLEU score, and perception quality along with font style preservation is required. Due to the absence of a robust automatic model that can verify cross-lingual font style similarity, we only consider Translation Quality and perception quality to compute combined $vt\text{-}score$ as follows:
\begin{equation}
    vt\text{-}score = \frac{2 \cdot TQ \cdot PQ}{TQ + PQ}.
\end{equation}
Please note that for images that contain only one word, we employ BLEU-1 instead of BLEU-2 to assess translation quality. For the remaining images, BLEU-2 is utilized in the above mentioned scoring measure.

\subsection{User Evaluation}
\label{sec:userEval}
Despite the availability of automatic evaluation metrics, as discussed above, user evaluation is crucial for assessing the accuracy, usability, and effectiveness of visual translations. User feedback is essential for evaluating the clarity, cultural appropriateness, and accessibility of translations. To this end, we conducted an extensive user evaluation with four human users aged 20 to 25 who hold graduate degrees and are proficient in both Hindi and English. They reviewed each visual translation baseline using Beamer slides: slides for metrics (ii) and (iii) featured single output images, while those for metrics (i) and (iv) displayed both input and output images together. The user evaluation metrics are described here:

\noindent\textbf{(i) Translation Quality (TQ) (score range: 1-4)}: This criterion focuses on the accuracy of the translation. Users were asked to rate whether the translated text accurately conveys the meaning of the original text. A higher score indicates a more accurate translation. The different ratings by users convey the following: 4: Linguistically and culturally totally correct translation. 3: Some words are correct; translation can be improved. 2: Very few words are correct, and significant improvement is required. 1: Totally incorrect translation. 

\noindent\textbf{(ii) Readability (R) (score range: 1-4)}: This criterion evaluates how easily the translated text can be read within the scene image. Factors such as font size, contrast, and placement of the text may influence readability. A higher score indicates better readability. The different readability ratings by users convey the following:
4: Clearly readable. 3: Can read with some effort. 2: Can read with significant effort; some words are not readable. 1: No text present in the target language.

\noindent\textbf{(iii) Perceptual Quality (PQ) (score range: 1-4)}: This criterion assesses how well the translated text blends into the scene image, making it difficult to distinguish from a real image. A higher score indicates better integration of the translated text with the scene. Users were asked to rate approaches based on the following: 4: Very clear, looks like real image. 3: Clear image, but some patches are present if carefully seen. 2: There are a lot of patchy effects; looks like a fake image. 1: Too much patchy effect; for sure, it is a fake image. 

\noindent\textbf{(iv) Source Style Preservation (SSP) (score range: 1-4)}: This criterion examines whether the translated text preserves the style, font, color, and other visual attributes of the original text in the scene image. A higher score indicates that the translated text maintains consistency with the source text in terms of visual presentation.
4: Font style, size, color, and background are coherent to the source. 3: Only 2 or 3 of the following: font style, size, color, and background are coherent to the source. 2: Only 1 or 2 of the following: font style, size, color, and background are coherent to the source. 1: No source-style preservation.

\begin{table*}[!t]
\centering
\scriptsize
\begin{tabular}
{llllccccc}
\hline
Method& STR     & MT     & STS    & {D.E.}  & TQ (BL-1) & TQ (BL-2)    & PQ & VT-score\\
\midrule
\multicolumn{9}{c}{\cellcolor[gray]{0.8}{\textbf{English-to-Hindi Translation}}}\\
\midrule
B-7  & DBNet+ParSeq  & Indic  & SRNet++  & \checkmark      &     \textbf{25.28}    & \textbf{20.54}    &53.79     & \textbf{27.51} \\

B-6  & DBNet+ParSeq  & Indic  & SRNet++  & \xmark       &     22.57 & 15.69     & 53.93    & 25.59 \\

B-5  & Oracle& Indic  & SRNet++  & \xmark        & 
22.36 & 16.90     & 53.38   &23.95\\

B-4  & DBNet+ParSeq  & M2M     & SRNet++ & \xmark       &     19.09    &  14.51    & \textbf{54.02}      &21.52 \\

B-3  & Oracle& M2M     & SRNet++  & \xmark        & 
19.82 &   15.33     & 53.52    & 22.22\\

B-2  & Oracle & M2M     & Mostel    & \xmark        & 
14.13 &  10.44      &  46.98    &16.58 \\

B-1  & Oracle & M2M    & SRNet     & \xmark        & 
15.00 & 12.25     & 46.71    & 16.56 \\
\midrule
\multicolumn{9}{c}{\cellcolor[gray]{0.8}{\textbf{Hindi-to-English Translation}}}\\
\midrule
B-7  & Oracle   & Indic  & SRNet++  & \checkmark      & \textbf{38.30}         & \textbf{29.30}         &    55.49         & \textbf{40.08}\\

B-6  & DBNet+ParSeq  & Indic  & SRNet++  & \xmark       &     29.10   &18.51  & \textbf{55.77}   & 28.52  \\

B-5  & Oracle  & Indic  & SRNet++  & \xmark        & 
31.31     & 19.70   & 55.62     & 32.27\\

B-4  & DBNet+ParSeq  & M2M     & SRNet++  & \xmark       &     03.22  & 02.19   & 55.60         & 03.81 \\

B-3  & Oracle  & M2M     & SRNet++  & \xmark        & 
04.20   & 02.89    & 55.58            &04.97  \\

B-2  & Oracle   & M2M     & Mostel    & \xmark        &02.03 
 &   01.40 &    53.41     & 02.46\\

B-1  & Oracle  & M2M     & SRNet     & \xmark        & 04.20
  &    02.86  &53.82   & 04.92 \\

\bottomrule
\end{tabular}
\caption{\textbf{Automatic Evaluation to evaluate baselines for visual translation.} We report translation quality (TQ) using BLEU-1 (BL-1) and BLEU-2 (BL-2) metrics and perception quality (PQ). D.E.: Design Enhancements. More details in Section~\ref{sec:autoEval}.}
\label{tab:auto}
\end{table*}

\section{Experiments}
\label{sec:experiments}
In this section, we comprehensively evaluate scene-text to scene-text translation baseline approaches discussed in Section~\ref{subsec:baselines} using both automatic and user evaluation metrics proposed in Section~\ref{sec:pm}. We use \realdata{} introduced in Section~\ref{sec:data} for all our evaluation. 

The automatic evaluation results are reported in Table~\ref{tab:auto}. We observe that SRNet++ clearly emerges as the best scene text synthesis approach as compared to other existing architectures. The proposed design enhancements also significantly boost translation quality while maintaining nearly identical perceptual quality.  We also observe that usage 
of IndicTrans2 as a translator consistently leads to an increase in translation quality. The state-of-the-art scene text recognition approaches are as good as ground truth annotations (Oracle) in the case of detecting and recognizing English text.

We further perform a rigorous user study using metrics presented in Section~\ref{sec:userEval}. As discussed in this section, we have collected user feedback from four qualified users and report mean scores of TQ, R, PQ, and SSP in Table~\ref{tab:user}. These scores nearly align with observations made via automatic evaluation. We also observe that there is significant room for improvement on all these metrics, indicating the challenge associated with the task.

\begin{table*}[!t]\
\scriptsize
\centering
\begin{tabular}{p{1cm}p{2.6cm}p{1cm}p{1.5cm}p{0.6cm}p{1 cm}p{1cm}p{1cm}p{0.8cm}}
\hline
Method &~STR & MT & STS & D.E. & TQ & R & PQ & SSP\\
\midrule
\multicolumn{9}{c}{\cellcolor[gray]{0.8}{\textbf{English-to-Hindi Visual Translation}}}\\
\midrule
B-7  & DBNet+ParSeq  & Indic  & SRNet++  & \checkmark      &       \textbf{2.25}  &    2.60 & 2.27        &    1.85\\

B-6  & DBNet+ParSeq  & Indic  & SRNet++  & \xmark       &   2.05      &   2.86   &  2.86       &   \textbf{1.97}  \\

B-5  & Oracle & Indic  & SRNet++  & \xmark       & 2.13 &    3.00 &    2.92     &    1.96  \\

B-4  & DBNet+ParSeq  & M2M     & SRNet++  & \xmark       & 1.93        &  3.12   &     \textbf{2.94 }    & 1.91     \\

B-3  & Oracle  & M2M     & SRNet++  & \xmark        & 1.94 & \textbf{3.27}  &   2.72     &  1.92     \\

B-2  & Oracle & M2M     & Mostel    & \xmark        &1.88  &   2.65&   2.42      &     1.85   \\

B-1  & Oracle  & M2M     & SRNet     & \xmark        & 1.94 &   2.51  &  2.50       & 1.88    \\
\midrule
\multicolumn{9}{c}{\cellcolor[gray]{0.8}{\textbf{Hindi-to-English Visual Translation}}}\\
\midrule
B-7  & Oracle  & Indic  & SRNet++  & \checkmark      &       \textbf{2.42}  &    \textbf{2.45}  & 2.19 & \textbf{1.79}   \\

B-6  & DBNet+ParSeq  & Indic  & SRNet++  & \xmark       &    1.92     &  2.11    & 2.05  &   1.67  \\

B-5  & Oracle  & Indic  & SRNet++  & \xmark       & 2.23   &  2.30 &\textbf{2.23} & 1.75     \\

B-4  & DBNet+ParSeq  & M2M     & SRNet++  & \xmark       &  1.36&   2.07  &1.95  & 1.42     \\

B-3  & Oracle   & M2M     & SRNet++  & \xmark    &1.64  &  2.19 &2.15 &  1.56     \\

B-2  & Oracle  & M2M     & Mostel    & \xmark        & 1.38 &  2.03 &1.94 &  1.63      \\

B-1  & Oracle    & M2M     & SRNet     & \xmark        & 1.53 &  2.09 & 1.96&   1.58\\
\bottomrule

\end{tabular}
\caption{\textbf{User Study to evaluate baselines for visual translation.} We report mean Translation Quality (TQ), Readability (R), Perception Quality (PQ), and Source Style Preservation (SSP). Four fluent Hindi-English speakers rated the output on a four-point Likert scale, with 4 being the highest quality. D.E.: Design Enhancements. For more details please refer to Section~\ref{sec:userEval}.}

\label{tab:user}
\end{table*}

\begin{figure*}[!t]
\centering
\includegraphics[width=0.9\textwidth]
{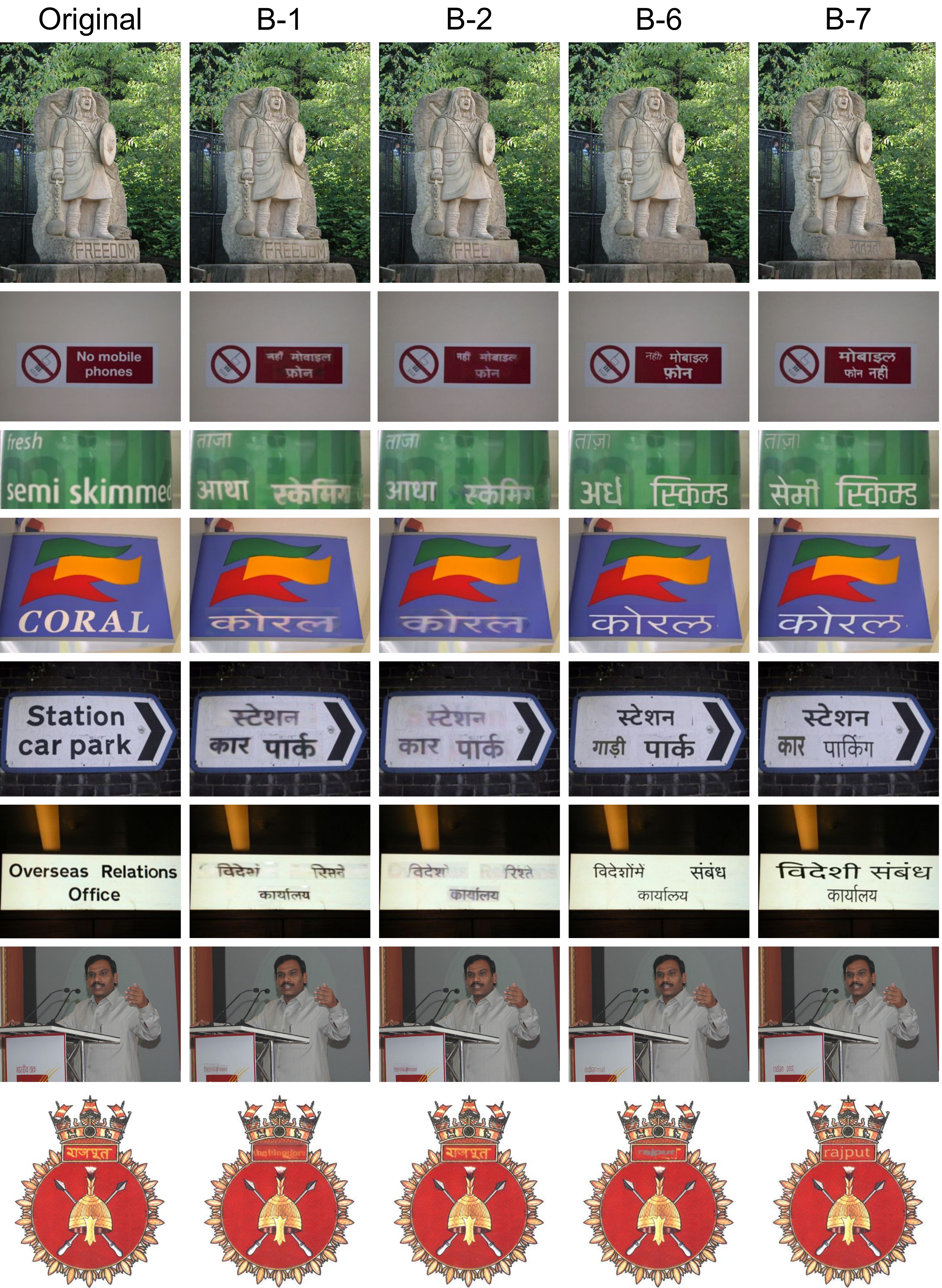}
\caption{A selection of visual translation results for proposed baseline variants for Eng $\rightarrow$ Hin in row 1--6 and Hin $\rightarrow$ Eng in row 7--8. Here, we show (left to right) the original image and results of best-performing baselines B-1, B-2, B-6, and B-7 (please refer to Section \ref{subsec:baselines} for details about these baseline variants). We observe that B-7, which uses SRnet++ for scene text synthesis and proposed design enhancements, is clearly superior in visual translation. Native Hindi speakers can find that IndicTrans2 (used in B-6 and B-7) produces superior translations, and the design enhancements in B-7 result in grammatically correct translations.}
\label{fig:qualitative-analysis}
\end{figure*}

\subsection{Qualitative Results}
We show a selection of visual results for the proposed baseline variants in Fig.~\ref{fig:qualitative-analysis}. The illustrated results indicate the merits/demerits of various choices. The use of IndicTrans2 as the translator instead of M2M improves translation to a large extent and also enables the transliteration of words when necessary. By using a ControlNet-based model for erasing scene-text regions in the image, in \textbf{SRNet++} instead of precariously erasing text from word crops as done by MOSTEL or SRNet, we ensure complete erasing of the source text. The rendering of the target text is also clearer and has a less patchy effect. \textbf{Design enhancements}, particularly translating at paragraph level instead of word level, improve the translation correctness by taking care of language-specific ordering of words.

\begin{figure*}[!t]
\centering
\includegraphics[width=0.5\textwidth]
{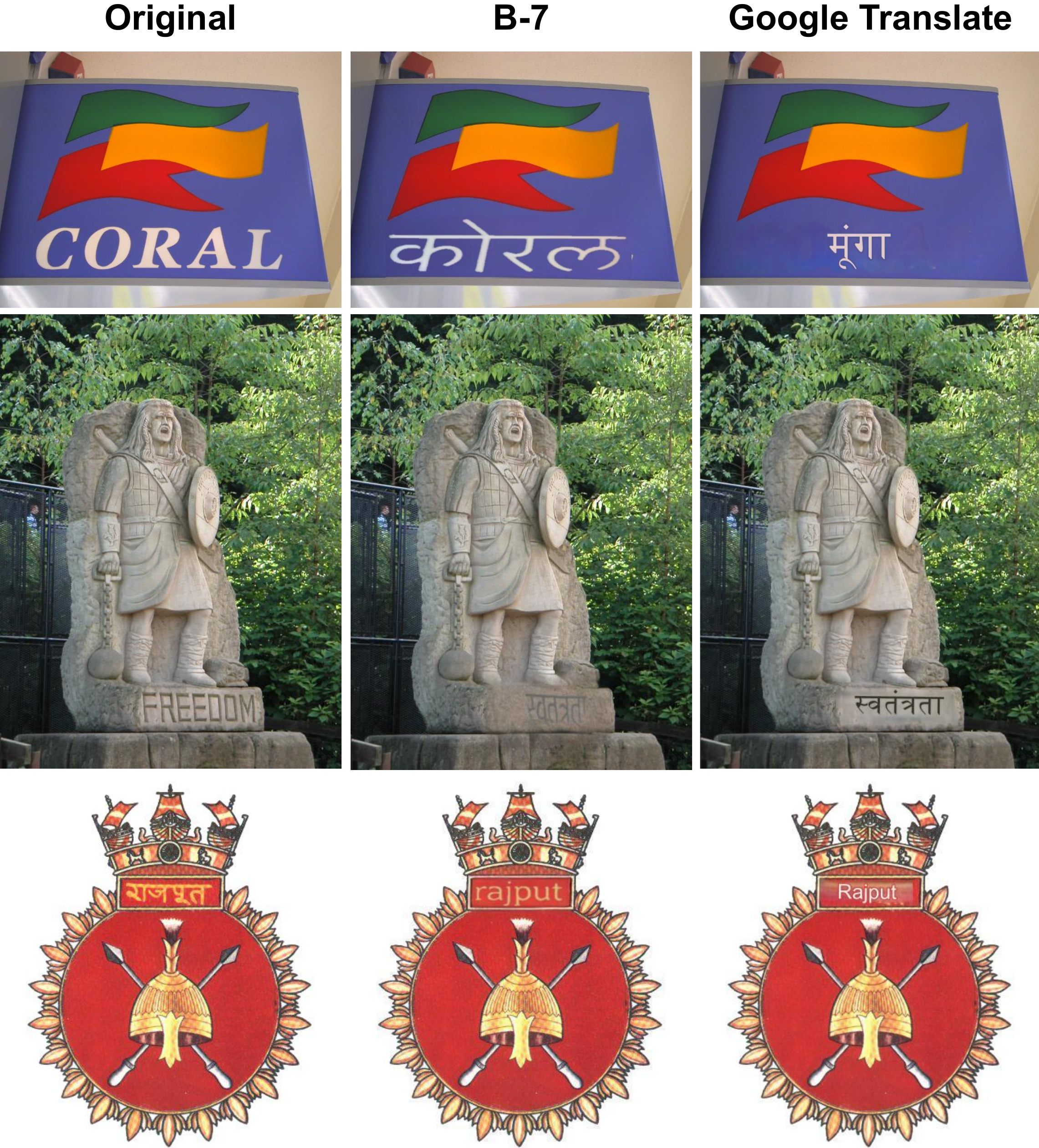}
\caption{Comparison of our proposed baseline (B-7) with Google Translate for images (a commercial application). For more details please refer to Section~\ref{sec:comm}. }
\label{fig: comparison-with-googletrans}
\end{figure*}

\subsection{Comparison with Commercial Systems}
\label{sec:comm}
Google Translate for images\footnote{\url{https://translate.google.com/?op=images}} is a commercial system that also handles scene text-to-scene text translation. However, it is closed-source and only available through a web interface, with no free API support. As a result, we do not include it in our quantitative comparison. Nevertheless, Fig.~\ref{fig: comparison-with-googletrans} presents some qualitative comparisons, illustrating that even Google Translate is not without flaws.

\subsection{Limitations}
The limitations of the proposed baselines are as follows: (i) It has limited success in visually translating curved or occluded Hindi texts, partly because, unlike English, scene text detection and recognition for Indian languages are still in their infancy. (ii) There is a trade-off between image and translation quality. Design enhancements allow for sentence-level translation, but approximations in word positioning and size can cause slight blurring, as illustrated in Fig.~\ref{fig:qualitative-analysis}. While these issues affect perceptual quality, we prioritize translation accuracy over image sharpness as long as the text remains readable. (iii) Ensuring the natural alignment of generated text in the scene is challenging. Therefore, the baselines have limited success in translating longer sentences or phrases. (iv) Our baselines do not utilize visual cues from the scene, which impairs their ability to choose between transliteration and translation, particularly for brand names. Additionally, the absence of an automatic metric for evaluating source style preservation or the visual consistency between source and generated scene text, such as font, orientation, and style, limits our current evaluation framework. Addressing these limitations is an important direction for future research.

\section{Conclusion}
We have presented a comprehensive study for the task of visual translation by proposing a series of baselines that utilize state-of-the-art approaches and their enhancements across various modules. Our baselines demonstrate promising results for translating scene text images between English and Hindi. However,  it is evident that visual translation remains challenging, and addressing all of its complexities extends beyond the scope of this single paper. We hope that introducing this task, along with the dataset, baseline, and performance metrics, will inspire the research community to develop advanced models for visual translation. 

\section*{Acknowledgement}
This work was partly supported by MeitY, Government of India under NLTM-Bhashini.

\scriptsize{
\bibliographystyle{splncs04}
\bibliography{references}}

\begin{thebibliography}{10}
\providecommand{\url}[1]{\texttt{#1}}
\providecommand{\urlprefix}{URL }
\providecommand{\doi}[1]{https://doi.org/#1}

\bibitem{BSTD}
{B}harat {S}cene {T}ext {D}ataset.
  \url{https://github.com/Bhashini-IITJ/BharatSceneTextDataset} (2024)

\bibitem{BahdanauCB14}
Bahdanau, D., Cho, K., Bengio, Y.: Neural machine translation by jointly
  learning to align and translate. In: ICLR (2015)

\bibitem{bautista2022scene}
Bautista, D., Atienza, R.: Scene text recognition with permuted autoregressive
  sequence models. In: ECCV (2022)

\bibitem{chen2014systematic}
Chen, B., Cherry, C.: A systematic comparison of smoothing techniques for
  sentence-level bleu. In: WMT@ACL (2014)

\bibitem{desai2014domain}
Desai, P., Sangodkar, A., P.~Damani, O.: A domain-restricted, rule based,
  english-hindi machine translation system based on dependency parsing. In:
  ICON (2014)

\bibitem{edunov2018understanding}
Edunov, S., Ott, M., Auli, M., Grangier, D.: Understanding back-translation at
  scale. In: EMNLP (2018)

\bibitem{fan2021beyond}
Fan, A., Bhosale, S., Schwenk, H., Ma, Z., El{-}Kishky, A., Goyal, S., Baines,
  M., Celebi, O., Wenzek, G., Chaudhary, V., Goyal, N., Birch, T., Liptchinsky,
  V., Edunov, S., Auli, M., Joulin, A.: Beyond {E}nglish-centric {M}ultilingual
  {M}achine {T}ranslation. J. Mach. Learn. Res.  \textbf{22},  107:1--107:48
  (2021)

\bibitem{gala2023indictrans2}
Gala, J.P., Chitale, P.A., AK, R., Gumma, V., Doddapaneni, S., M., A.K.,
  Nawale, J.A., Sujatha, A., Puduppully, R., Raghavan, V., Kumar, P., Khapra,
  M.M., Dabre, R., Kunchukuttan, A.: {I}ndic{T}rans2: Towards high-quality and
  accessible machine translation models for all 22 scheduled indian languages.
  Trans. Mach. Learn. Res.  \textbf{2023} (2023)

\bibitem{gehring2017convolutional}
Gehring, J., Auli, M., Grangier, D., Yarats, D., Dauphin, Y.N.: Convolutional
  sequence to sequence learning. In: ICML (2017)

\bibitem{hurskainen2017rule}
Hurskainen, A., Tiedemann, J.: Rule-based machine translation from english to
  finnish. In: WMT (2017)

\bibitem{jia2019direct}
Jia, Y., Weiss, R.J., Biadsy, F., Macherey, W., Johnson, M., Chen, Z., Wu, Y.:
  Direct speech-to-speech translation with a sequence-to-sequence model. In:
  Interspeech (2019)

\bibitem{kano2021transformer}
Kano, T., Sakti, S., Nakamura, S.: Transformer-based direct speech-to-speech
  translation with transcoder. In: IEEE Spoken Language Technology Workshop
  (SLT) (2021)

\bibitem{Karatzas2013ICDAR2R}
Karatzas, D., Shafait, F., Uchida, S., Iwamura, M., i~Bigorda, L.G., Mestre,
  S.R., Mas, J., Mota, D.F., Almaz{\'{a}}n, J., de~las Heras, L.: {ICDAR} 2013
  robust reading competition. In: ICDAR (2013)

\bibitem{krishnan2023textstylebrush}
Krishnan, P., Kovvuri, R., Pang, G., Vassilev, B., Hassner, T.: Textstylebrush:
  Transfer of text aesthetics from a single example. {IEEE} Trans. Pattern
  Anal. Mach. Intell.  \textbf{45}(7),  9122--9134 (2023)

\bibitem{lan2024translatotron}
Lan, Z., Niu, L., Meng, F., Zhou, J., Zhang, M., Su, J.: Translatotron-v
  (ison): An end-to-end model for in-image machine translation. In: ACL
  (Findings) (2024)

\bibitem{lan2023exploring}
Lan, Z., Yu, J., Li, X., Zhang, W., Luan, J., Wang, B., Huang, D., Su, J.:
  Exploring better text image translation with multimodal codebook. In: ACL
  (2023)

\bibitem{lee2022direct}
Lee, A., Chen, P.J., Wang, C., Gu, J., Popuri, S., Ma, X., Polyak, A., Adi, Y.,
  He, Q., Tang, Y., et~al.: Direct speech-to-speech translation with discrete
  units. In: ACL (2022)

\bibitem{lee2021rewritenet}
Lee, J., Kim, Y., Kim, S., Yim, M., Shin, S., Lee, G., Park, S.: Rewritenet:
  Reliable scene text editing with implicit decomposition of text contents and
  styles. In: CVPRW (2022)

\bibitem{liao2020real}
Liao, M., Wan, Z., Yao, C., Chen, K., Bai, X.: Real-time scene text detection
  with differentiable binarization. In: AAAI (2020)

\bibitem{ma2022improving}
Ma, C., Zhang, Y., Tu, M., Han, X., Wu, L., Zhao, Y., Zhou, Y.: Improving
  end-to-end text image translation from the auxiliary text translation task.
  In: ICPR (2022)

\bibitem{Contrique}
Madhusudana, P.C., Birkbeck, N., Wang, Y., Adsumilli, B., Bovik, A.C.: Image
  quality assessment using contrastive learning. {IEEE} Trans. Image Process.
  \textbf{31},  4149--4161 (2022)

\bibitem{mansimov2020towards}
Mansimov, E., Stern, M., Chen, M.X., Firat, O., Uszkoreit, J., Jain, P.:
  Towards end-to-end in-image neural machine translation. In: EMNLP Workshop
  (2020)

\bibitem{mathew2017benchmarking}
Mathew, M., Jain, M., Jawahar, C.: Benchmarking scene text recognition in
  devanagari, telugu and malayalam. In: ICDAR (2017)

\bibitem{nakamura2006atr}
Nakamura, S., Markov, K., Nakaiwa, H., Kikui, G., Kawai, H., Jitsuhiro, T.,
  Zhang, J., Yamamoto, H., Sumita, E., Yamamoto, S.: The {ATR} multilingual
  speech-to-speech translation system. {IEEE} Trans. Speech Audio Process.
  \textbf{14}(2),  365--376 (2006)

\bibitem{otsu1975threshold}
Otsu, N.: A threshold selection method from gray-level histograms. {IEEE}
  Trans. Syst. Man Cybern.  \textbf{9}(1),  62--66 (1979)

\bibitem{pirinen2019apertium}
Pirinen, T.A.: Apertium-fin-eng--rule-based shallow machine translation for wmt
  2019 shared task. In: WMT (2019)

\bibitem{qu2023exploring}
Qu, Y., Tan, Q., Xie, H., Xu, J., Wang, Y., Zhang, Y.: Exploring stroke-level
  modifications for scene text editing. In: AAAI (2023)

\bibitem{roy2020stefann}
Roy, P., Bhattacharya, S., Ghosh, S., Pal, U.: {STEFANN}: scene text editor
  using font adaptive neural network. In: CVPR (2020)

\bibitem{DiffSceneTextEraser}
Susladkar, O.: {D}iff-{S}cene {T}ext {E}raser.
  \url{https://github.com/Onkarsus13/Diff_SceneTextEraser} (2023)

\bibitem{vaswani2017attention}
Vaswani, A., Shazeer, N., Parmar, N., Uszkoreit, J., Jones, L., Gomez, A.N.,
  Kaiser, L., Polosukhin, I.: Attention is all you need. In: NeurIPS (2017)

\bibitem{wu2019editing}
Wu, L., Zhang, C., Liu, J., Han, J., Liu, J., Ding, E., Bai, X.: Editing text
  in the wild. In: ACM-MM (2019)

\bibitem{yang2020swaptext}
Yang, Q., Huang, J., Lin, W.: Swaptext: Image based texts transfer in scenes.
  In: CVPR (2020)

\bibitem{controlNet}
Zhang, L., Rao, A., Agrawala, M.: Adding conditional control to text-to-image
  diffusion models. In: ICCV (2023)

\bibitem{zhao2019muse}
Zhao, G., Sun, X., Xu, J., Zhang, Z., Luo, L.: Muse: Parallel multi-scale
  attention for sequence to sequence learning. arXiv preprint arXiv:1911.09483
  (2019)

\end{thebibliography}

\end{document}